\documentclass[letterpaper, 10pt, conference]{ieeeconf}

\IEEEoverridecommandlockouts   
\usepackage{placeins}
\usepackage{float}
\usepackage[utf8]{inputenc}
\usepackage[T1]{fontenc}
\usepackage{microtype}

\usepackage{amsmath,amsfonts,amssymb}
\usepackage{stmaryrd}
\usepackage{nicefrac}

\usepackage{algorithm}
\usepackage{algorithmic}

\usepackage{graphicx}
\usepackage{array}
\usepackage{booktabs}
\usepackage{stfloats}
\usepackage[caption=false,font=footnotesize]{subfig}

\usepackage{textcomp}
\usepackage[table]{xcolor}
\usepackage[normalem]{ulem}

\usepackage{xurl}
\usepackage[bookmarks=true,hidelinks]{hyperref}
\usepackage[capitalise,nameinlink]{cleveref}

\crefname{figure}{Fig.}{Figs.}
\Crefname{figure}{Fig.}{Figs.}
\crefname{section}{Section}{Sections}
\crefformat{equation}{(#2#1#3)}
\Crefformat{equation}{(#2#1#3)}

\usepackage{etoolbox}
\apptocmd{\thebibliography}{\emergencystretch=2em\relax}{}{}

\newif\ifarxiv
\arxivtrue  

\title{Amortising Trajectory Optimisation for Residual MPC via Implicit Contact Differentiation}

\ifarxiv
\author{
    Daniel Layeghi\textsuperscript{1}\textsuperscript{*},
    Thomas Corb\`{e}res\textsuperscript{1}\textsuperscript{*},
    Calum Arnott\textsuperscript{1}\textsuperscript{\dag},
    Aditya Kamireddypalli\textsuperscript{1},\\
    Hashim Al-Obaidi\textsuperscript{1},
    Steve Tonneau\textsuperscript{1},
    Michael Mistry\textsuperscript{1}
      \thanks{\textsuperscript{1}School of Informatics, University of Edinburgh, Scotland.}%
      \thanks{\textsuperscript{*}These authors contributed equally.}%
      \thanks{\textsuperscript{\dag}Corresponding author: \texttt{c.r.arnott@ed.ac.uk}.}%
  }
\else
  \author{
    \thanks{Submitted for double-anonymous review; author identities and affiliations omitted.}}
\fi

\newcommand{\vc}[1]{\mathbf{#1}} 					

\def\@onedot{\ifx\@let@token.\else.\null\fi\xspace}

\newcommand{\Expect}{{\rm I\kern-.3em E}}				

\begin{document}
\newcommand{\stn}[2][red]{\textcolor{#1}{#2}}
\newcommand{\stb}[2][red]{\textcolor{#1}{\sout{#2}}}
\newcommand{\str}[2]{\textcolor{red}{\sout{#1}{2}}}

\maketitle

\begin{abstract}
Differentiable simulation can accelerate contact-rich trajectory optimisation by exposing local sensitivities of task outcomes to controls. Existing approaches either use finite differences, which are expensive and step-size sensitive; differentiate iterative contact solvers by unrolling automatic differentiation (AD), which stores a growing computation trace; or require intricate, solver-specific KKT sensitivity derivations. We introduce an AD-assisted implicit derivative for regularised smooth contacts and apply it to Mujoco MJX, based on the Implicit Function Theorem (IFT). The method differentiates the stationarity residual at the tolerance-converged solution, avoiding both solver unrolling and hand-assembled KKT systems. IFT keeps compiled temporary memory nearly constant with solver effort, changing by less than 4\% from one to ten iterations versus 10.6× growth for unrolled AD. IFT memory grows slower with active contacts and model dimension, using 20× less memory at 256 contacts and 6× less at 16 contacts and 96 DoF. We further introduce optimiser distillation for residual MPC, amortising batched full-horizon iLQR into a policy that guides short-horizon residual iLQR. Across Finger, Franka, and Unitree, this raises six-step success by 28-98 percentage points over standard iLQR. 
\end{abstract}

\begingroup
\newcommand{\widthFig}{0.495\columnwidth}  
\newcommand{\heightFig}{0.25\columnwidth}  
\setlength{\tabcolsep}{0.005\columnwidth} 

\newcommand{\subFigA}[1]{%
  \raisebox{-0.5\height}{%
    \includegraphics[clip,trim=350px 250px 350px 215px,width=\widthFig,height=\heightFig]{#1}%
  }%
}
\newcommand{\subFigB}[1]{%
  \raisebox{-0.5\height}{%
    \includegraphics[clip,trim=280px 220px 200px 100px,width=\widthFig,height=\heightFig]{#1}%
  }%
}
\newcommand{\subFigC}[1]{%
  \raisebox{-0.5\height}{%
    \includegraphics[clip,trim=280px 300px 200px 20px,width=\widthFig,height=\heightFig]{#1}%
  }%
}
\newcommand{\subFigD}[1]{%
  \raisebox{-0.5\height}{%
    \includegraphics[clip,trim=300px 300px 300px 100px,width=\widthFig,height=\heightFig]{#1}%
  }%
}
\begin{figure}[t]
  \centering
  \begin{tabular}{@{}cc@{}}
    \subFigA{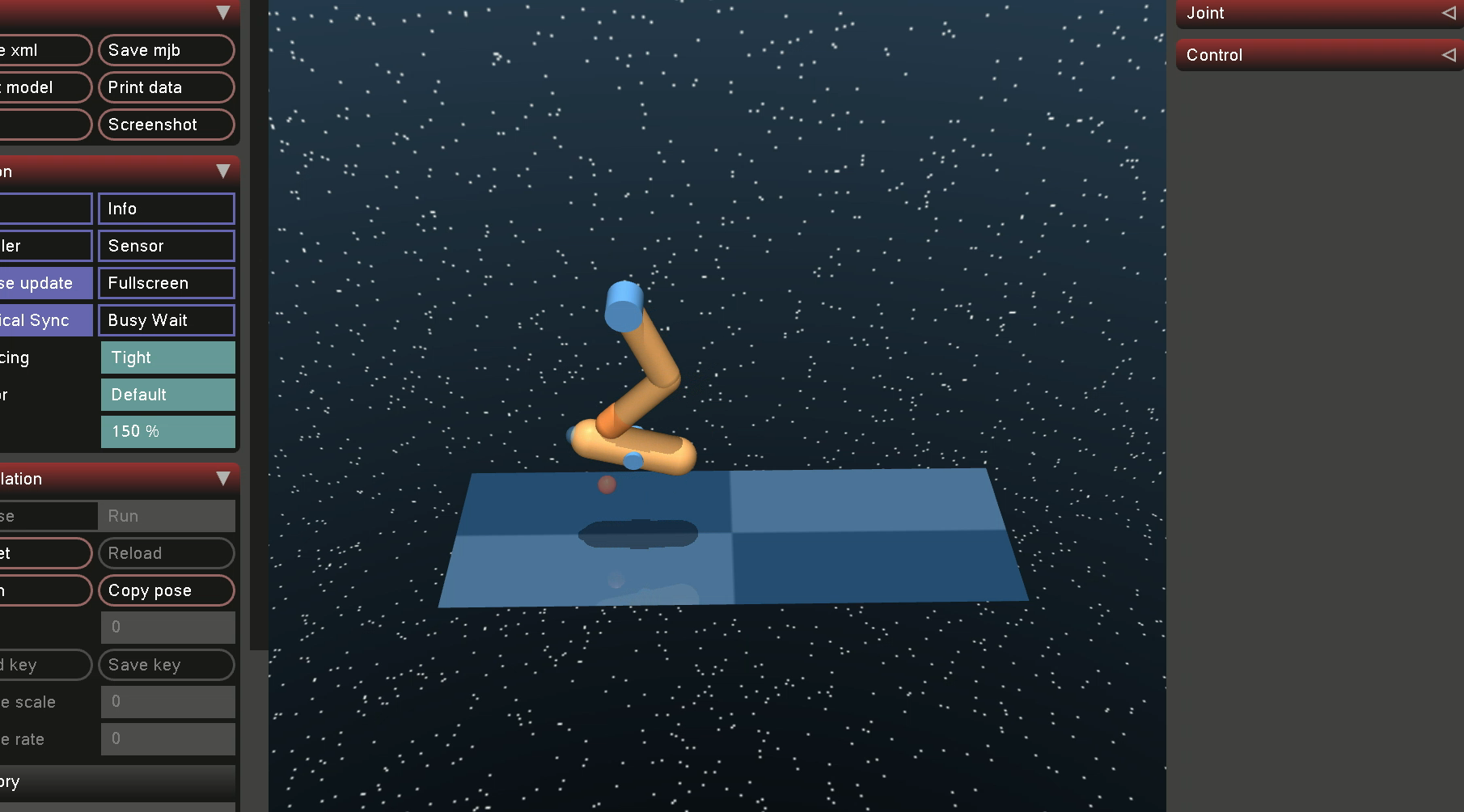} &
    \subFigB{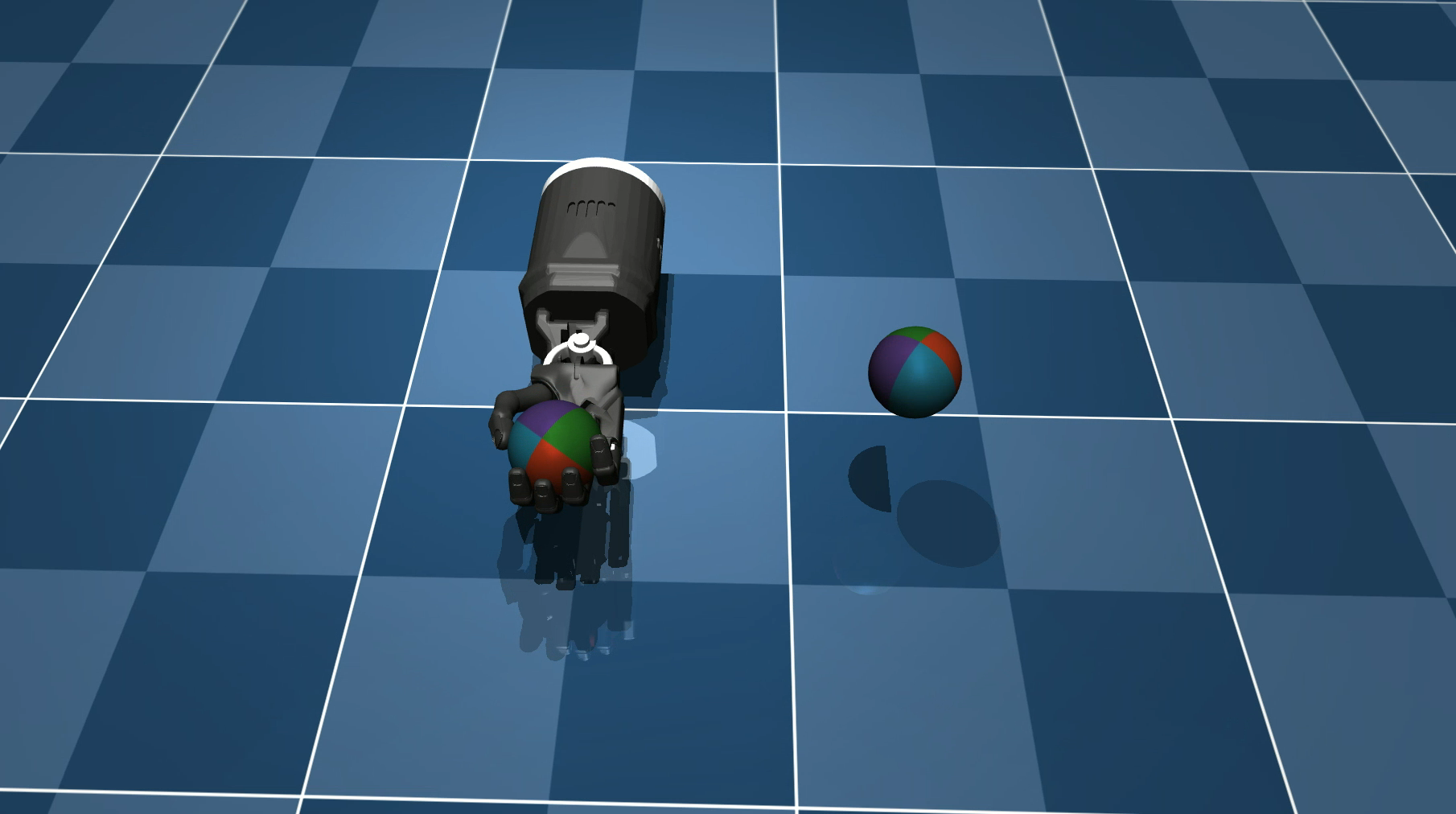} \\
    \\ [-2.15ex]
    \subFigC{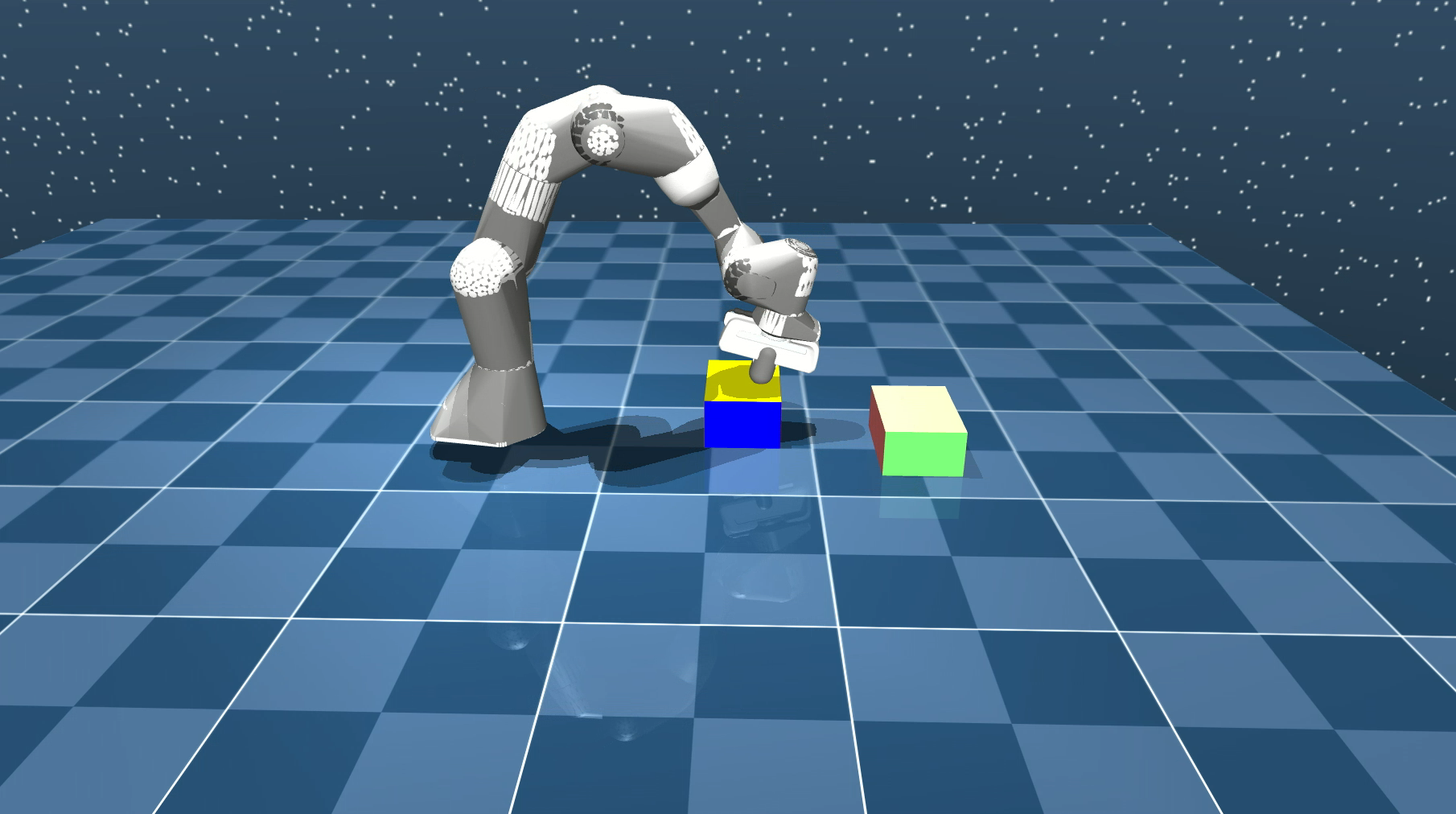} &
    \subFigD{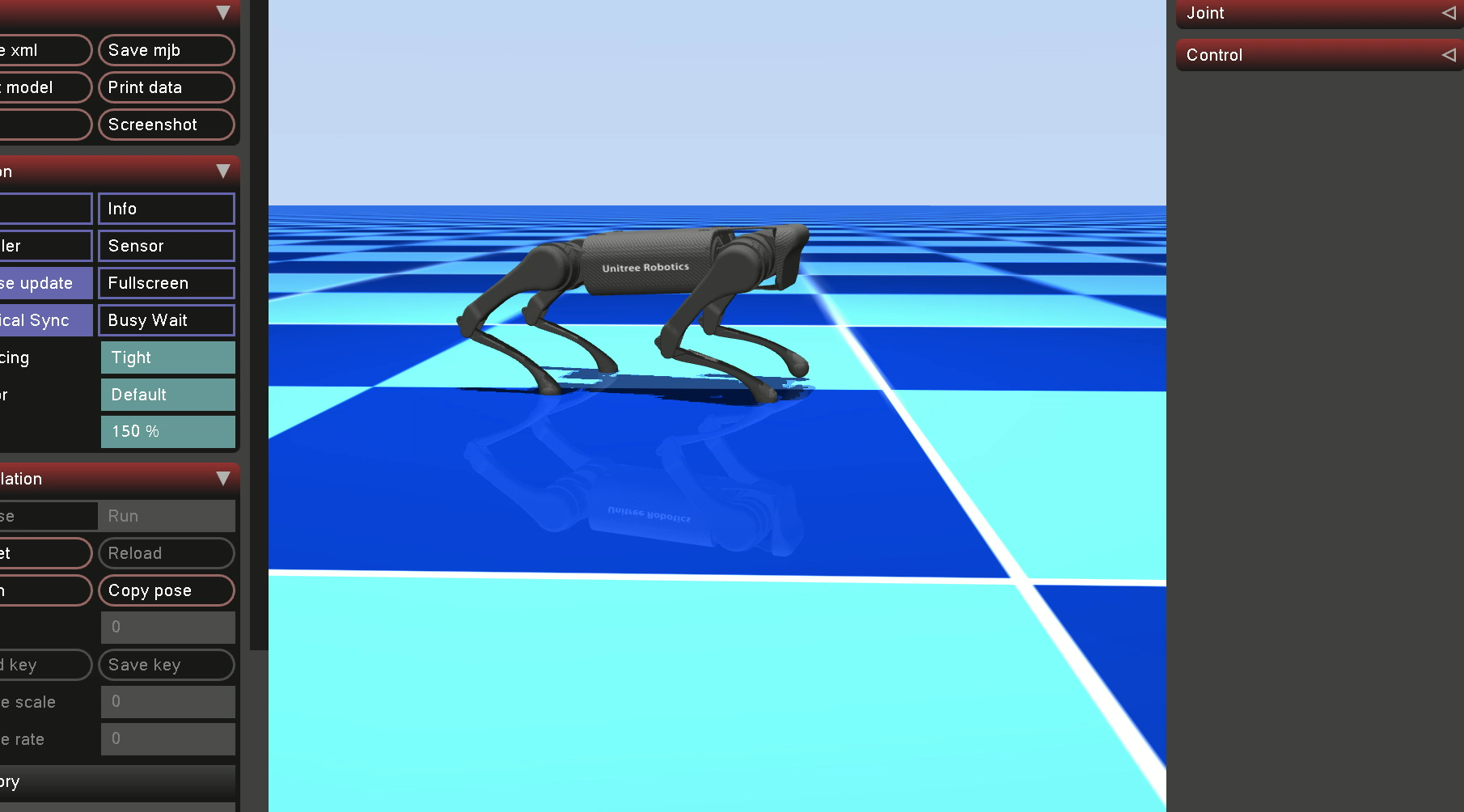}
  \end{tabular}
  \caption{Examples of trajectories solved across different scenarios with our framework.}
  \label{fig:scenario-comparison}
\end{figure}
\endgroup

\section{Introduction}
\label{sec:introduction}
Differentiable simulation connects contact-rich planning and learning. Local derivatives show how controls affect future task outcomes, enabling trajectory optimisation. In many cases optimised trajectories are used to supervise policies, amortising expensive task-specific computation across initial conditions and goals.

Scaling this pipeline is difficult because contact is commonly resolved by an iterative constrained solve. Under standard unrolled reverse-mode automatic differentiation (AD), every solver iteration becomes part of the backward trace for each contacted step and trajectory. Tightening the contact solution therefore consumes more memory and leaves room for fewer parallel trajectories; truncating it saves memory but differentiates an under-solved contact response. Rematerialisation reduces stored intermediates by replaying the (K)-step solver during the backward pass, so its computation still grows with solver effort. This creates a convergence--parallelism trade-off. 

We address this trade-off with a residual-based implicit derivative for regularised smooth contact. Using the Implicit Function Theorem (IFT), we derive the local sensitivity of the tolerance-converged contact solution and instantiate it in Mujoco MJX without changing the forward solver. We compare its gradient agreement and memory scaling with finite differences and unrolled AD as solver effort, active contacts, and generalized velocity dimension increase. The derivative is local to fixed smooth contact regions and may change across contact-set or friction-mode transitions.

We then use this scalability to introduce optimizer distillation for residual MPC. Batched full-horizon iLQR generates and continually refreshes contact-rich teacher trajectories, whose optimized actions are distilled into a policy. At deployment, the policy provides the long-horizon nominal action, while short-horizon residual iLQR supplies local model-based corrections. We evaluate how this amortization changes the success rate and planning-horizon trade-off relative to standard iLQR on Finger, Franka, and Unitree.

\noindent\textbf{Contributions.} \textbf{(1)} We introduce an AD-assisted implicit backward pass for the stationarity residual of MJX's regularised contact solve. It leaves the forward solver unchanged, avoids hand-assembled KKT sensitivities, and removes solver-iteration-dependent reverse-mode storage. \textbf{(2)} We validate whole-rollout gradients against central finite differences, showing that IFT matches AD and FD in accuracy at high iteration counts while substantially outperforming them in compiled temporary memory across solver iterations, active contacts, and generalised velocity dimension. We use the resulting derivatives in batched full-horizon trajectory optimisation. \textbf{(3)} We distil full-horizon optimised actions into a policy paired with short-horizon residual iLQR, and evaluate closed-loop behaviour on Finger, Franka, and Unitree. We release the implementation as open source.\footnote{\ifarxiv\url{https://github.com/calumarnott/mujoco}\else \url{https://anonymous.4open.science/r/mujoco-ift/} \fi}

\section{Background}
\label{sec:background}

A contact-rich simulator step is often defined by an iterative numerical solve rather than an explicit transition map. Its derivative therefore depends on whether one differentiates the executed solver trace or the converged solution. We review three areas relevant to this distinction: modular implicit differentiation, differentiable contact simulation and batched trajectory optimisation, and optimiser distillation for policy-guided MPC.

\subsection{Implicit differentiation as a modular solver interface}

Optimisation layers differentiate solutions through optimality conditions rather than through the iterations used to compute them. OptNet differentiates quadratic programs through their Karush--Kuhn--Tucker (KKT) system, while differentiable convex-programming layers canonicalise broader problem classes to cone programs and differentiate the resulting solution map
\cite{amos2017optnet,agrawal2019differentiable}. These methods are general within their supported problem classes, but their backward passes require an explicitly represented optimisation problem.

Modular implicit-differentiation frameworks instead separate the forward solver from its derivative. Blondel et al.\ \cite{blondel2022efficient} use a user-defined root, fixed-point, or optimality map and apply automatic differentiation to form the Jacobian products required by the Implicit Function Theorem (IFT). Related work constructs matrix-free implicit derivatives for energy-based soft-body simulation using reverse-mode AD
\cite{rojas2021differentiable}. We apply similar modular pattern to continuos and differentiable contact simulation of rigid bodies and instantiate it in Mujoco/MJX.

\subsection{Differentiable contact and batched trajectory optimisation}

Rigid contact is modelled through complementarity conditions, conic constraints, or regularised convex programs
\cite{stewart1996implicit,anitescu2006optimization,todorov2014convex}. These formulations differ in their physical approximations and numerical smoothness. Regularisation can give a unique, locally smooth solution within a fixed contact set and smooth frictional regime, but it does not remove non-smoothness at contact creation, removal, or friction-mode changes.

Differentiable simulators also differ in how they construct contact sensitivities. Finite differences repeat the simulator evaluation in each input direction. Belbute-Peres et al.\ \cite{belbuteperes2018endtoend} differentiate the KKT system of an LCP recast as a quadratic program; Nimble derives sparse analytical LCP sensitivities
\cite{werling2021fast}; and Dojo implicitly differentiates a custom primal--dual nonlinear-complementarity solver
\cite{howell2022dojo}. Other work uses localised implicit programs or analytical collision and friction derivatives
\cite{qiao2020scalable,lelidec2024endtoend}. These methods construct the contact backward pass from solver-specific analytical, KKT, or residual structure.

Differentiable-programming simulators instead expose their implemented operations to AD and accelerator batching
\cite{hu2020difftaichi,freeman2021brax,deepmind2026mjx}. When contact is solved iteratively, standard reverse mode differentiates the finite solver trace and stores its intermediate states. This storage issue is distinct from the quality of gradients near nonsmooth events such as contact creation, changing normals, impacts, stiff regularisation, or missing contacts and is part of what we address in this paper.
\cite{zhong2022differentiable,zhong2023improving}. DiffMJX addresses the derivative accuracy through adaptive integration and a backward-only contact-from-distance mechanism
\cite{paulus2025hard}. Our method addresses the truncation error caused by low solver iteration by using IFT to remove the memory scaling bottleneck without changing the forward contact model or the integration method.

Trajectory optimisation adds a further scaling requirement. Contact-implicit methods can discover contact-rich behaviour
\cite{posa2014direct,kurtz2022contactimplicit}, while iLQR/DDP and sparse optimal-control libraries exploit local dynamics structure
\cite{tassa2012synthesis,mastalli2020crocoddyl}. Finite-difference and approximated dynamics derivatives have also supported demanding control
\cite{zhang2026wholebody,russell2023adaptive}. Batched and GPU-parallel trajectory optimisation is well established
\cite{rastgar2021gpu,amatucci2025primaldual,du2025gato}; our concern is the memory cost of the contact derivative within those batches. With unrolled reverse-mode AD, every contacted step stores the intermediate state of each solver iteration, so tighter solves leave less memory for parallel full-horizon trajectories. This also affects offline distillation, where teacher trajectories are generated in batches. We quantify this dependence in \cref{sec:method}.

\subsection{Optimiser distillation and policy-guided MPC}

Optimiser distillation uses optimised trajectories to train a policy. Guided Policy Search, PLATO, and MPC-Net amortise trajectory-centric or MPC teachers into policies that replace online planning
\cite{levine2016endtoend,kahn2017plato,carius2020mpcnet}. The distillation loss itself does not require simulator derivatives. Differentiable MPC takes a different approach: it differentiates the fixed-point KKT conditions of an MPC problem to learn costs or dynamics end-to-end
\cite{amos2018differentiablempc}. Action distillation instead treats optimised controls as supervised targets and need not differentiate through the trajectory optimiser.

Hybrid methods retain both learning and planning at deployment. POPLIN uses a policy to initialise or parametrise derivative-free planning over a learned model
\cite{wang2019poplin}, while learned networks have also been used to warm-start convex MPC problems
\cite{chen2022largescale}. Residual policy learning adds a learned correction to a fixed model-based controller
\cite{johannink2019residual}. We reverse this assignment: the distilled policy supplies the long-horizon nominal action, while a model-based optimiser computes a short-horizon residual correction and local feedback gains through the differentiable contact model.

Other methods shorten online planning by amortising long-horizon information into a terminal value or planning prior. POLO learns a terminal-value ensemble for short-horizon sampling-based MPC
\cite{lowrey2019polo}, while TD-MPC2 plans in a learned latent model using both a terminal value and a policy prior
\cite{hansen2024tdmpc2}. Our policy instead provides the nominal action about which a residual and its feedback gains are optimised. Terminal-value methods are complementary and could use the same full-horizon teacher trajectories.

SHAC learns policies directly through differentiable simulation using short differentiation horizons to improve contact-gradient quality
\cite{xu2022accelerated}. Contact-implicit MPC has also reached long-horizon and hierarchical contact-rich control without amortisation
\cite{jiang2024cimpc,suh2025ctr,xie2026wheretotouch}, typically using solver-specific or hand-assembled contact sensitivities. We combine a modular residual derivative for MJX contact, batched full-horizon iLQR, and optimiser-in-the-loop distillation. To isolate the effect of policy guidance, we compare the resulting residual controller with standard short-horizon iLQR under the same differentiable model in Section~\ref{sec::learning-results}.

\section{Implicit Function Theorem}
\label{sec:method}

We seek derivatives of the converged MJX contact solution without storing
or traversing the iterations used to compute it. The key is to
differentiate a stationarity residual that vanishes at the solution.
Strong convex regularisation makes the solved acceleration unique. Within
a fixed contact set and a smooth cone-projection region, the residual is
continuously differentiable and has a nonsingular Jacobian with respect
to acceleration. The resulting derivative is local: it need not remain
valid across contact creation, contact removal, or friction-mode changes.

We retain MuJoCo's round Coulomb cone rather than replacing it with a
polyhedral approximation. Below, we introduce only the contact objective,
its stationarity residual, and the corresponding implicit backward rule.
Further derivations and comparisons of contact formulations are given in
\cite{todorov2012mujoco,lidec2024contact}, and MJX implements the
regularised model used here \cite{deepmind2026mjx}.

\subsection{Regularised Cone Contact Dynamics}
\label{subsec:regularised_contact}

Let $\vc q\in\mathbb R^{n_Q}$, $\vc v\in\mathbb R^{n_V}$, and
$\vc a=\dot{\vc v}\in\mathbb R^{n_V}$ denote configuration, generalised
velocity, and generalised acceleration. With mass matrix
$\vc M(\vc q)\succ0$, generalised bias force $\vc c(\vc q,\vc v)$,
actuation $\boldsymbol\tau$, contact Jacobian $\vc J$, and contact force
$\boldsymbol\lambda$, the unconstrained and constrained accelerations
satisfy
\begin{equation}
\label{eq:rigid_body_dynamics}
    \vc a_{\mathrm{unc}}
    =
    \vc M^{-1}(\boldsymbol\tau-\vc c),
    \qquad
    \vc M\vc a+\vc c
    =
    \boldsymbol\tau+\vc J^\mathsf T\boldsymbol\lambda .
\end{equation}

\paragraph{Admissible contact forces.}
For exposition, consider contacts with one normal and two tangential
force components. At contact $i$, write
\[
    \boldsymbol\lambda_i
    =
    \begin{bmatrix}
        \lambda_{N,i}\\
        \boldsymbol\lambda_{T,i}
    \end{bmatrix}
    \in
    \mathbb R\times\mathbb R^2.
\]
Contact can push but not pull, so $\lambda_{N,i}\geq0$. Coulomb friction
bounds the tangential force by $\mu_i\lambda_{N,i}$, where $\mu_i\geq0$
is the friction coefficient. The admissible force cone is
\begin{equation}
\label{eq:local_coulomb_cone}
    \mathcal K_i
    =
    \left\{
        \boldsymbol\lambda_i
        \in\mathbb R\times\mathbb R^2
        :
        \lambda_{N,i}\geq0,\;
        \|\boldsymbol\lambda_{T,i}\|_2
        \leq
        \mu_i\lambda_{N,i}
    \right\}.
\end{equation}

Its dual cone is
\begin{equation}
\label{eq:dual_cone}
    \mathcal K^\ast
    =
    \left\{
        \vc z\in\mathbb R^m:
        \boldsymbol\lambda^\mathsf T\vc z\geq0
        \quad
        \text{for every }
        \boldsymbol\lambda\in\mathcal K
    \right\}.
\end{equation}
Thus, $\mathcal K$ describes admissible contact forces, while
$\mathcal K^\ast$ describes compatible contact-space residuals.

MuJoCo embeds this cone geometry in a regularised convex problem. Let
$\vc a_{\mathrm{ref}}\in\mathbb R^m$ denote the reference acceleration
constructed from the current constraint state and solver parameters, and
let the diagonal matrix $\vc R\succ0$ set the contact regularisation.
Introducing a contact-space variable $\vc y\in\mathbb R^m$, the
frictional contact block is
\begin{equation}
\label{eq:ccp_primal_full}
\begin{aligned}
    (\vc a^\star,\vc y^\star)
    =
    \operatorname*{arg\,min}_{\vc a,\vc y}
    \quad&
    \frac{1}{2}
    \|\vc a-\vc a_{\mathrm{unc}}\|_{\vc M}^{2}
    +
    \frac{1}{2}
    \|\vc y-\vc a_{\mathrm{ref}}\|_{\vc R^{-1}}^{2}
    \\
    \mathrm{s.t.}\quad&
    \vc J\vc a-\vc y\in\mathcal K^\ast .
\end{aligned}
\end{equation}
The first term penalises deviation from unconstrained motion, while the
second determines the compliant contact response. The feasible set is
convex, and the objective is strongly convex because
$\vc M\succ0$ and $\vc R^{-1}\succ0$. The minimiser is therefore unique.
For clarity, we show only the frictional contact block; other regularised
constraint rows contribute analogous terms.

For fixed $\vc a$, define
$\vc u\triangleq\vc J\vc a-\vc a_{\mathrm{ref}}$ and substitute
$\vc z=\vc J\vc a-\vc y$. The inner minimisation becomes a weighted
projection onto $\mathcal K^\ast$. Following MuJoCo
\cite{todorov2012mujoco}, define
\begin{equation}
\label{eq:ccp_primal_reduced}
\begin{aligned}
    s(\vc u;\vc\theta)
    &\triangleq
    \min_{\vc z\in\mathcal K^\ast(\vc\theta)}
    \frac{1}{2}
    \|\vc u-\vc z\|_{\vc R^{-1}}^2,
    \\
    \mathcal L(\vc a;\vc\theta)
    &\triangleq
    \frac{1}{2}
    \|\vc a-\vc a_{\mathrm{unc}}\|_{\vc M}^{2}
    +
    s\!\left(
        \vc J\vc a-\vc a_{\mathrm{ref}};
        \vc\theta
    \right),
    \\
    \vc a^\star(\vc\theta)
    &=
    \operatorname*{arg\,min}_{\vc a}
    \mathcal L(\vc a;\vc\theta),
\end{aligned}
\end{equation}
where
$\vc\theta=(\vc q,\vc v,\boldsymbol\tau,\boldsymbol\mu,\vc R,\ldots)$
collects the quantities entering the contact step.

Because $\mathcal K^\ast$ is nonempty, closed, and convex and
$\vc R\succ0$, the projection defining $s$ is unique. Consequently,
$s$ is convex and continuously differentiable in $\vc u$, and
$-\nabla_{\vc u}s$ gives the regularised contact force. The strongly
convex inertial term then makes the reduced objective strongly convex, so
\eqref{eq:ccp_primal_reduced} has the same unique acceleration
$\vc a^\star$ as \eqref{eq:ccp_primal_full}. Within a fixed contact set
and smooth projection region, the stationarity residual of this reduced
problem is differentiable. For full derivation of Mujoco contact model refer to \cite{todorov2012mujoco}.

\subsection{Stationarity Residual and Iterative Solution}
\label{subsec:solver_fixed_point}

The reduced objective generally has no closed-form minimiser because
$s$ contains a projection onto the friction cone. MJX therefore solves
the problem iteratively using Newton or nonlinear conjugate-gradient
methods. Both methods target the same unique stationary point; the
implicit backward rule depends on that stationary point rather than on
the particular iterations used to reach it.

Define the stationarity residual
\begin{equation}
\label{eq:ccp_stationarity}
\begin{aligned}
    \vc F(\vc a,\vc\theta)
    &=
    \nabla_{\vc a}
    \mathcal L(\vc a;\vc\theta)
    \\
    &=
    \vc M
    \left(
        \vc a-\vc a_{\mathrm{unc}}
    \right)
    +
    \vc J^\mathsf T
    \nabla_{\vc u}s
    \left(
        \vc J\vc a-\vc a_{\mathrm{ref}};
        \vc\theta
    \right).
\end{aligned}
\end{equation}
The exact minimiser is the unique root
\begin{equation}
\label{eq:stationarity_root}
    \vc F(\vc a^\star,\vc\theta)=\vc0.
\end{equation}
Thus, $\vc F$ is both the gradient of the reduced objective and the
generalised force imbalance at the solution.

For the implemented MJX solve, the same residual can be evaluated as
\begin{equation}
\label{eq:mjx_stationarity}
    \vc F
    =
    \vc M\vc a
    -
    \vc f_{\mathrm{smooth}}
    -
    \vc f_{\mathrm{constraint}},
\end{equation}
where $\vc f_{\mathrm{smooth}}$ contains the smooth generalised forces
and $\vc f_{\mathrm{constraint}}$ contains the generalised force produced
by the regularised constraints. Automatic differentiation can therefore
differentiate $\vc F$ directly through the implemented residual
computation.

To connect this residual to the forward solver, consider the Newton
method used in our experiments. Within a fixed contact set and smooth
projection region,
\begin{equation}
\label{eq:newton_matrix}
\begin{aligned}
    \vc B(\vc a,\vc\theta)
    &=
    \frac{\partial\vc F}{\partial\vc a}
    \\
    &=
    \nabla_{\vc a}^{2}
    \mathcal L(\vc a;\vc\theta)
    \\
    &=
    \vc M
    +
    \vc J^\mathsf T
    \nabla_{\vc u}^{2}s
    \left(
        \vc J\vc a-\vc a_{\mathrm{ref}};
        \vc\theta
    \right)
    \vc J.
\end{aligned}
\end{equation}
Because $s$ is convex,
$\nabla_{\vc u}^{2}s\succeq0$ wherever the Hessian exists. Since
$\vc M\succ0$,
\[
    \vc B\succ0,
\]
so the local Newton system is nonsingular.

At iteration $k$, Newton's method computes
\[
    \vc p_k
    =
    -\vc B_k^{-1}\vc F_k,
    \qquad
    \vc B_k
    =
    \vc B(\vc a_k,\vc\theta),
    \qquad
    \vc F_k
    =
    \vc F(\vc a_k,\vc\theta).
\]
A line search selects a positive step length $\alpha_k$ and applies
\begin{equation}
\label{eq:newton_update}
\begin{aligned}
    \vc a_{k+1}
    &=
    \vc T(\vc a_k,\vc\theta)
    \\
    &=
    \vc a_k+\alpha_k\vc p_k
    \\
    &=
    \vc a_k
    -
    \underbrace{
        \alpha_k\vc B_k^{-1}
    }_{\displaystyle \vc P_k}
    \vc F(\vc a_k,\vc\theta).
\end{aligned}
\end{equation}
The matrix $\vc P_k$ combines the Newton metric and line-search scale.
Locally nonsingular damping or preconditioning can be absorbed into the
same matrix. On a neighbourhood where the accepted solver branch is fixed,
the update can therefore be written as
\begin{equation}
\label{eq:solver_map}
    \vc T(\vc a,\vc\theta)
    =
    \vc a
    -
    \vc P(\vc a,\vc\theta)
    \vc F(\vc a,\vc\theta).
\end{equation}

Since $\vc F(\vc a^\star,\vc\theta)=\vc0$, the stationary solution is
also a fixed point:
\[
    \vc T(\vc a^\star,\vc\theta)=\vc a^\star.
\]
Define the iteration-map residual
\begin{equation}
\label{eq:iteration_residual}
\begin{aligned}
    \vc G(\vc a,\vc\theta)
    &=
    \vc a-\vc T(\vc a,\vc\theta)
    \\
    &=
    \vc P(\vc a,\vc\theta)
    \vc F(\vc a,\vc\theta).
\end{aligned}
\end{equation}
If $\vc P$ is nonsingular, then
$\vc G(\vc a,\vc\theta)=\vc0$ if and only if
$\vc F(\vc a,\vc\theta)=\vc0$. The two residuals therefore identify the
same local solution.

They also produce the same local solution derivative. Let
\[
    \vc P^\star
    =
    \vc P(\vc a^\star,\vc\theta).
\]
Differentiating $\vc G=\vc P\vc F$ at the solution gives
\begin{equation}
\label{eq:residual_equivalence}
    \vc G_{\vc a}
    =
    \vc P^\star\vc F_{\vc a},
    \qquad
    \vc G_{\vc\theta}
    =
    \vc P^\star\vc F_{\vc\theta}.
\end{equation}
Terms containing derivatives of $\vc P$ vanish because they are
multiplied by
$\vc F(\vc a^\star,\vc\theta)=\vc0$. Hence,
\begin{equation}
\label{eq:equal_root_sensitivity}
\begin{aligned}
    -
    \vc G_{\vc a}^{-1}\vc G_{\vc\theta}
    &=
    -
    \left(
        \vc P^\star\vc F_{\vc a}
    \right)^{-1}
    \vc P^\star\vc F_{\vc\theta}
    \\
    &=
    -
    \vc F_{\vc a}^{-1}\vc F_{\vc\theta}.
\end{aligned}
\end{equation}
The Newton metric and line search therefore affect how the forward solver
reaches the root, but not the local derivative of the isolated root.
For this derivative, the backward pass need not traverse the Newton
iterates or differentiate the line search, warm start, or stopping logic,
provided the forward solver reaches the same local solution.

\subsection{Implicit Differentiation of the Contact Solution}
\label{subsec:implicit_differentiation}

For each parameter vector $\vc\theta$, let
$\vc a^\star(\vc\theta)$ denote the unique exact contact solution. Its
defining relation is
\begin{equation}
\label{eq:implicit_solution}
    \vc F
    \left(
        \vc a^\star(\vc\theta),
        \vc\theta
    \right)
    =
    \vc0.
\end{equation}
MJX approximates this root to its configured tolerance, subject to a
maximum-iteration safeguard.

Within a fixed contact set and smooth projection region, $\vc F$ is
continuously differentiable. By \eqref{eq:newton_matrix},
\[
    \vc F_{\vc a}
    \triangleq
    \frac{\partial\vc F}{\partial\vc a}
    =
    \vc B
    \succ0,
\]
so $\vc F_{\vc a}$ is nonsingular. The Implicit Function Theorem gives
\begin{equation}
\label{eq:ift_sensitivity}
    \vc F_{\vc a}
    \frac{\mathrm d\vc a^\star}{\mathrm d\vc\theta}
    +
    \vc F_{\vc\theta}
    =
    \vc0,
    \qquad
    \frac{\mathrm d\vc a^\star}{\mathrm d\vc\theta}
    =
    -
    \vc F_{\vc a}^{-1}
    \vc F_{\vc\theta},
\end{equation}
where all Jacobians are evaluated at
$(\vc a^\star,\vc\theta)$. This is the derivative of the local contact
solution defined by the root condition, rather than the derivative of a
finite solver trace.

For a scalar loss
$\ell(\vc a^\star(\vc\theta),\vc\theta)$, the chain rule gives
\[
    \frac{\mathrm d\ell}{\mathrm d\vc\theta}
    =
    \nabla_{\vc\theta}\ell
    -
    \nabla_{\vc a}\ell^\mathsf T
    \vc F_{\vc a}^{-1}
    \vc F_{\vc\theta}.
\]
There is no need to form either the full sensitivity matrix or the
inverse. Instead, define the adjoint
$\boldsymbol\eta\in\mathbb R^{n_V}$ by
\begin{equation}
\label{eq:ift_reverse}
    \vc F_{\vc a}^{\mathsf T}\boldsymbol\eta
    =
    \nabla_{\vc a}\ell,
    \qquad
    \frac{\mathrm d\ell}{\mathrm d\vc\theta}
    =
    \nabla_{\vc\theta}\ell
    -
    \boldsymbol\eta^\mathsf T\vc F_{\vc\theta}.
\end{equation}
Reverse mode therefore requires one transposed linear solve and one
vector--Jacobian product, rather than explicitly forming
$\mathrm d\vc a^\star/\mathrm d\vc\theta$.

We implement \eqref{eq:ift_reverse} with
\texttt{jax.lax.custom\_root}, leaving the MJX forward solver unchanged.
Automatic differentiation provides derivatives of the stationarity
residual, and a dense QR solve applies \(\vc F_a^{-\top}\) without relying
on exact numerical symmetry of the AD-constructed Jacobian. The resulting
cotangent is then propagated through integration, kinematics, and the
remaining simulator operations by standard automatic differentiation.
The adjoint cost depends on the generalised velocity dimension, but its
stored state and backward traversal do not grow with the number of
iterations $K$ used by the forward contact solver. No separately derived
primal--dual KKT backward system is required.

\subsection{IFT Results}
\label{sec:ift-results}

\begin{figure}[!t]
    \centering
    \includegraphics[width=\columnwidth]
        {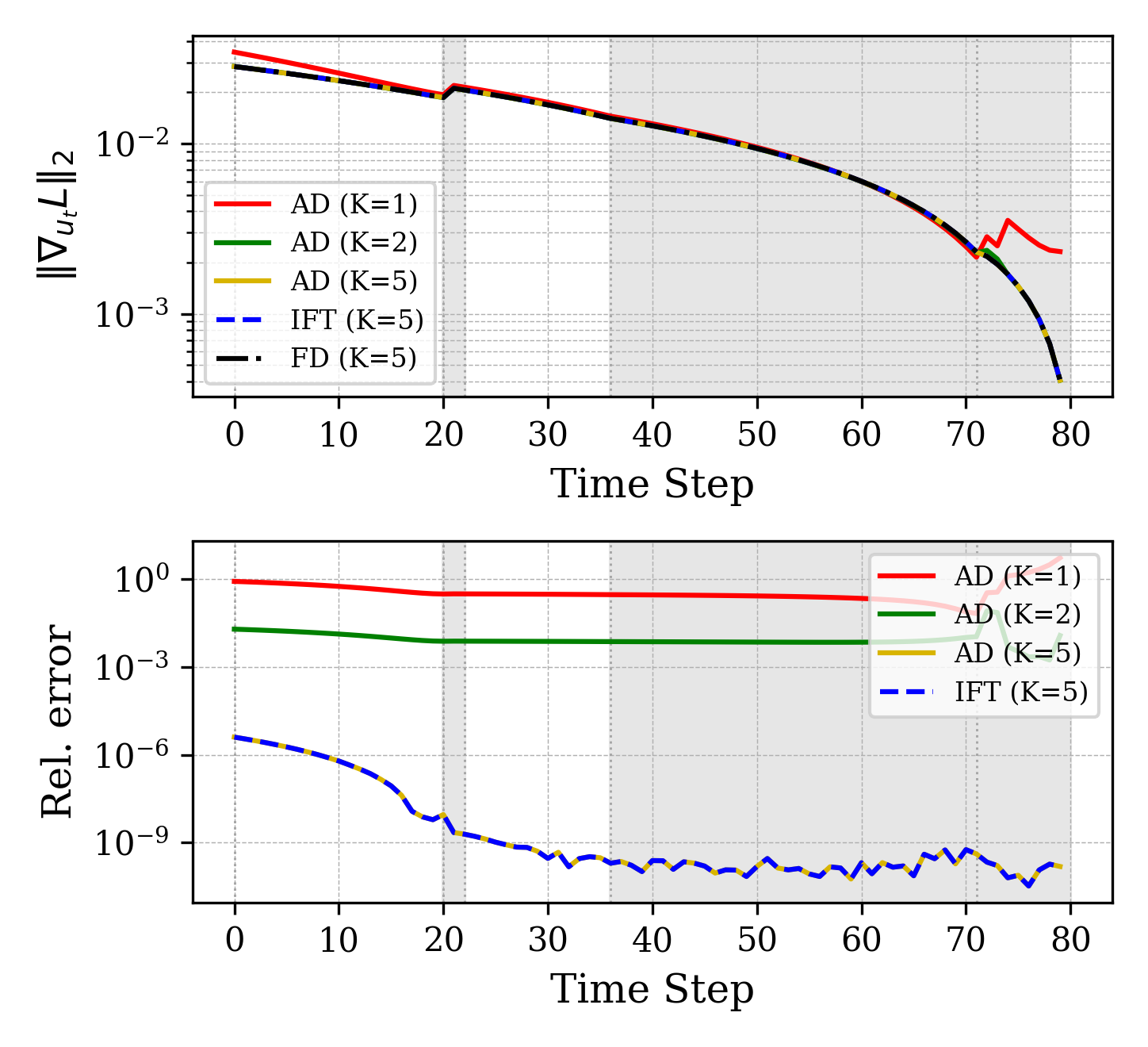}
    \caption{Open-loop sensitivity of a ball bounce-to-target rollout.
    Top: control-gradient magnitude over the horizon.
    Bottom: per-step relative discrepancy from whole-rollout central FD
    using the same \(K=5\) forward-solver configuration. The \(K=5\)
    unrolled-AD curve overlaps the IFT curve in the lower panel.
    Shading marks active-contact intervals, and dotted lines mark
    contact-set changes. All 480 FD perturbations preserve the nominal
    contact schedule.}
    \label{fig:ball-gradient}
\end{figure}

\begin{figure*}[!t]
    \centering
    \includegraphics[width=0.99\textwidth]
        {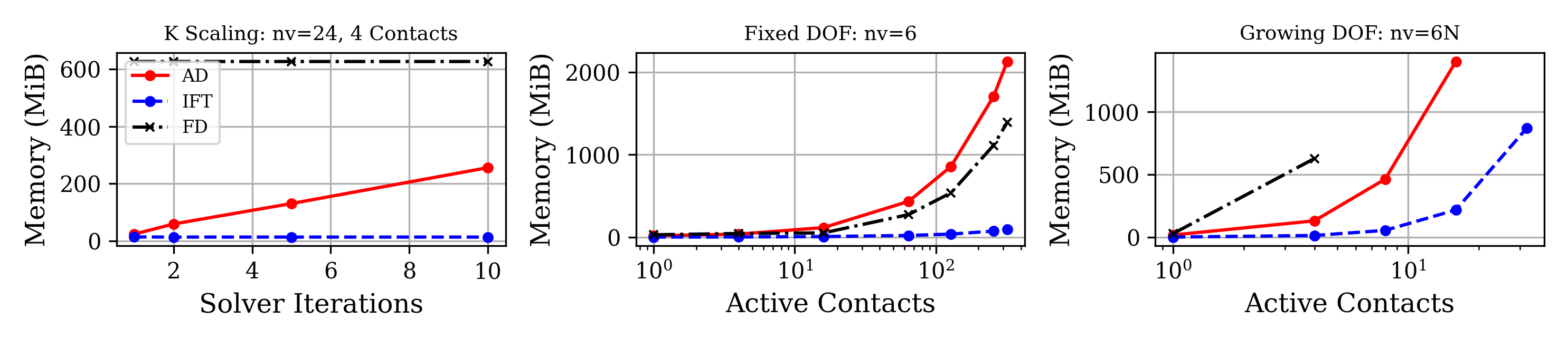}
    \caption{Compiled temporary memory for batched differentiation through
    primitive-contact steps at \(B=1024\).
    Left: solver-iteration scaling for four particles with \(n_V=24\)
    and four active contacts.
    Centre: the number of active contacts increases on one free body while
    \(n_V=6\) remains fixed.
    Right: \(N\) free particles each contribute one active contact, giving
    \(n_V=6N\).
    Missing points denote GPU out-of-memory failures. The centre and right
    panels use \(K=5\).}
    \label{fig:memory-scaling}
\end{figure*}

We compare residual-based IFT, unrolled reverse-mode AD, and finite
differences (FD), using double precision throughout. Unrolled AD
differentiates the executed \(K\)-iteration solver trace, whereas IFT
differentiates the stationarity residual at the tolerance-converged
forward solution. Figure~\ref{fig:ball-gradient} uses whole-rollout
central FD to validate the resulting sensitivities.
Figure~\ref{fig:memory-scaling} uses a separate parallel per-step FD
implementation only to study memory scaling. For the \(K=5\)
comparison, all methods use the same forward-solver configuration and
differ only in how the sensitivity is obtained.

\paragraph{Rollout-gradient validation}
We use a fixed \(H=80\) open-loop rollout of one free sphere launched
toward a floor and wall. The timestep is \(0.01\,\mathrm{s}\), the
nominal controls are zero with \(\vc u_t\in\mathbb R^3\), and the loss
penalises terminal position and velocity. The rollout contains free
flight, floor impact and separation, sustained floor contact, and a
later floor--wall interaction; 46 steps invoke contact. At the nominal
control, the terminal-loss contribution to an earlier control gradient
propagates through the complete rollout due to chain rule. Therefore, local errors in a contact derivative can affect controls well
before the corresponding contact event.

Figure~\ref{fig:ball-gradient} compares unrolled AD and IFT with
whole-rollout central FD. At \(K=5\), unrolled AD and IFT agree with FD
throughout the horizon, including during active contact. This agreement
provides an independent numerical check that the \(K=5\) trace
derivative is correct and has reached the same local sensitivity as the
implicit residual derivative. In contrast, the shorter \(K=1\) and
\(K=2\) unrolls retain errors concentrated around contact.

An \(\epsilon\)-sweep confirms a stable FD reference, and every
perturbation preserves the nominal contact schedule. Finite-\(K\) AD is
exact for its executed trace; for the shorter unrolls, that trace is
under-solved and therefore differs from the locally converged contact
map recovered by IFT and approached by AD as \(K\) increases.

\paragraph{Temporary-memory scaling}
Figure~\ref{fig:memory-scaling} reports compiled temporary memory for batched MJX contact steps on a
4\,GiB GPU. The left panel isolates solver effort at fixed model
dimension and contact count. Unrolled-AD memory grows with \(K\), while
IFT remains nearly constant because it stores no solver-iteration trace.

The centre panel increases the number of active contacts while holding
\(n_V\) fixed. Both methods incur the growing contact workspace, but IFT
remains substantially smaller. The right panel increases contacts and
model dimension together. Dense QR storage then produces approximately
quadratic IFT memory growth in \(n_V\), yet IFT completes the largest
tested problem, whereas unrolled AD and parallel FD run out of memory
earlier. Thus, IFT removes the \(K\)-dependent trace term; the ordinary
costs of batching, contact workspace, and model dimension remain.
\section{Trajectory Optimisation}\label{sec:trajopt}

\begin{figure*}[ht]
    \centering
    \includegraphics[width=\linewidth, keepaspectratio=true]{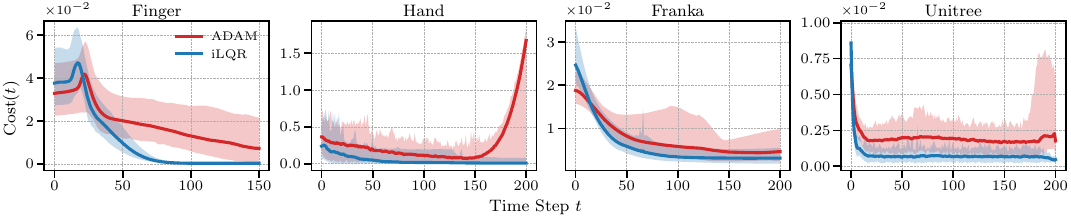}
    \vspace{-0.5cm}
    \caption{Running-cost profiles of the final trajectories produced by Adam and iLQR on four contact-rich tasks. Curves show the mean over 100 trajectories after 20 optimisation iterations; shaded regions show the 10th--90th percentile range.}
    \label{fig::comparison-ilqr-adam}
\end{figure*}

We next use the IFT-enabled MJX dynamics for batched trajectory
optimisation on the contact-rich tasks in
\cref{fig:scenario-comparison}. Removing the solver-iteration trace
from each contacted step preserves more batch capacity as the forward
contact solve is tightened. We compare Adam, which optimises the full
control sequence using first-order gradients, with iLQR, which exploits
the temporal structure of the control problem.

\subsection{Problem Formulation}

We consider finite-horizon trajectory optimisation with discrete-time
dynamics
\[
    \vc x_{t+1}
    =
    \vc f(\vc x_t,\vc u_t),
    \qquad
    t=0,\ldots,N-1,
\]
where \(\vc f\) denotes one MJX simulation step, including the
regularised contact solve. Given the initial state
\(\vc x_0=\vc x_{\mathrm{init}}\), we optimise the control sequence
\(U=(\vc u_0,\ldots,\vc u_{N-1})\) by solving
\begin{equation}
\label{eq:ocp}
\begin{aligned}
\min_{U}\quad
&
J(U)
=
\sum_{t=0}^{N-1}
\ell(\vc x_t,\vc u_t)
+
\ell_T(\vc x_N)
\\
\text{s.t.}\quad
&
\vc x_{t+1}
=
\vc f(\vc x_t,\vc u_t),
\qquad
\vc x_0
=
\vc x_{\mathrm{init}} .
\end{aligned}
\end{equation}

Both methods use the same objective and batched MJX dynamics.
Trajectory instances are vectorised across the batch, and each contact
solve within \(\vc f\) is differentiated using the implicit backward
pass from \cref{sec:ift-results}.

\subsection{First-Order Method: Adam}

Adam treats the complete control sequence \(U\) as the decision
variable and updates it using gradients obtained by reverse-mode
differentiation through the rollout. At optimisation iteration \(k\),
the update is written schematically as
\[
    U^{(k+1)}
    =
    \mathrm{Adam}
    \left(
        U^{(k)},
        \nabla_U J\!\left(U^{(k)}\right)
    \right).
\]

Differentiating through a long rollout requires storing intermediate
trajectory states. We therefore checkpoint the rollout, trading
recomputation for memory. Contact steps still use the implicit backward
pass and do not store the internal solver iterations.

\subsection{Structured Shooting Method: iLQR}

iLQR exploits the local structure of the dynamics and
cost-to-go~\cite{tassa2012synthesis,mastalli2020crocoddyl}. Around a
nominal trajectory \((\vc x_t,\vc u_t)\), it linearises the dynamics as
\[
    \delta\vc x_{t+1}
    =
    \vc A_t\,\delta\vc x_t
    +
    \vc B_t\,\delta\vc u_t,
    \quad
    \vc A_t
    =
    \frac{\partial \vc f}{\partial \vc x},
    \quad
    \vc B_t
    =
    \frac{\partial \vc f}{\partial \vc u}.
\]
The Jacobians \(\vc A_t\) and \(\vc B_t\) include the implicit contact
derivatives from \cref{sec:ift-results}.

Using a quadratic model of the cost-to-go, the standard Riccati
backward recursion computes feedforward and feedback gains
\(\vc k_t\) and \(\vc K_t\)~\cite{tassa2012synthesis}. We implement this
recursion as a backward scan over time steps. This horizon-level scan is
separate from the iterations of the contact solver, whose derivatives
are handled implicitly. The line search is batched over both the trajectory and step-size dimensions, so each trajectory selects its own step size.

\subsection{TO Results}

\Cref{fig::comparison-ilqr-adam} compares the trajectories produced by
Adam and iLQR after 20 optimisation iterations on 100 trajectories per
task. iLQR achieves lower mean running cost and tighter
10th--90th percentile ranges on all four tasks, with the clearest
separation on Hand and Unitree. We therefore use full-horizon iLQR to
generate the teacher trajectories used for policy distillation.
\section{Optimiser Distillation for Short-Horizon Contact MPC}
\label{sec::learning-results}

We amortise full-horizon contact optimisation into a policy for
short-horizon control. Full-horizon iLQR generates successful teacher
trajectories; at deployment, the policy supplies long-horizon nominal
actions and residual iLQR provides contact-aware local feedback. We
evaluate three tasks with randomised initial conditions: \textbf{Finger},
where a 2-DoF arm rotates a spinner to a target angle; \textbf{Franka},
where a 7-DoF arm pushes a box to a fixed target region; and
\textbf{Unitree}, where an A1 quadruped accelerates from rest toward
\(1\,\mathrm{m/s}\).

\subsection{Hybrid residual iLQR}
\label{sec::Hybrid-iLQR}

Let \(\vc o_t\) denote the policy observation, comprising the state and task information such as the goal. The applied control is
\[
\vc u_t
=
\operatorname{sg}\!\left[
    \boldsymbol\pi_{\bar\theta}(\vc o_t)
\right]
+
\vc r_t ,
\]
where \(\boldsymbol\pi_{\bar\theta}\) is the target policy, \(\vc r_t\) is the iLQR-optimised residual, and \(\operatorname{sg}[\cdot]\) denotes stop-gradient. The policy action remains in the forward rollout, but iLQR treats it as constant and optimises only the residual.

For a planning horizon \(H\), the hybrid controller solves
\begin{equation}
\begin{aligned}
\min_{\vc r_{0:H-1}} \quad
&
\sum_{t=0}^{H-1}
\ell_{\mathrm{hyb}}
\left(
    \vc x_t,\vc r_t;\vc x^\star
\right)
+
\ell_T(\vc x_H;\vc x^\star)
\\
\text{s.t.}\quad
&
\vc x_{t+1}
=
\vc f
\left(
    \vc x_t,
    \operatorname{sg}
    \left[
        \boldsymbol\pi_{\bar\theta}(\vc o_t)
    \right]
    +
    \vc r_t
\right),
\qquad
\vc x_0=\vc x_{\mathrm{init}} .
\end{aligned}
\label{eq:hybrid-ilqr-objective}
\end{equation}

The hybrid running cost is
\[
\ell_{\mathrm{hyb}}
\left(
    \vc x_t,\vc r_t;\vc x^\star
\right)
=
\ell_{\mathrm{task}}
\left(
    \vc x_t;\vc x^\star
\right)
+
\|\vc r_t\|_{\vc R}^{2},
\]
where \(\ell_{\mathrm{task}}\) is the task cost and \(\vc R\succ0\) penalises the residual. Penalising \(\vc r_t\), rather than the full action, keeps the optimised control close to the policy while allowing contact-aware local corrections.

\subsection{Online optimiser distillation}

Training uses full-horizon hybrid iLQR as an online teacher, as
summarised in Algorithm~1. Each collection batch samples independent
initial conditions and goals, and only successful teacher trajectories
are retained. For each retained trajectory, the supervised target is
the complete optimised action
\[
\vc u_t^\star
=
\mathrm{sg}\!\left[\boldsymbol\pi_{\bar{\theta}}(\vc o_t)\right]
+
\vc r_t^\star,
\]
where \(r_t^\star\) is the residual produced by full-horizon iLQR. The
resulting observation--action pairs are stored in replay buffer
\(\mathcal{B}\), and the policy is trained on mini-batches
\(\mathcal{M}\subset\mathcal{B}\) using
\begin{equation}
\mathcal{L}_{\mathrm{sup}}(\theta)
=
\frac{1}{|\mathcal{M}|}
\sum_{(\vc o,\vc u^\star)\in\mathcal{M}}
\left\lVert \boldsymbol\pi_\theta(\vc o)-\vc u^\star \right\rVert_2^2 .
\label{eq:supervised_loss} 
\end{equation}
Imitating the complete action distils the combined policy--optimiser
behaviour while retaining residual MPC for local correction. The target
policy is updated by exponential averaging,
\[
\bar{\theta}
\leftarrow
(1-\tau)\bar{\theta}+\tau\theta,
\]
which prevents abrupt policy changes from shifting the optimiser's
rollout distribution. Training uses full-horizon iLQR for supervision,
whereas evaluation uses short-horizon receding-horizon control.

\begin{algorithm}[th]
\algsetup{linenosize=\small}
\caption{\small Online optimiser distillation with hybrid iLQR}
\label{alg::online-training}
\begin{algorithmic}[1]
\small
\STATE Initialise policy \(\boldsymbol\pi_\theta\), target policy \(\boldsymbol\pi_{\bar\theta}\), and replay buffer \(\mathcal B\)
\FOR{each epoch}
    \STATE Sample a batch of initial conditions and goals
    \STATE Solve full-horizon hybrid iLQR using \(\boldsymbol\pi_{\bar\theta}\)
    \STATE Store successful pairs \((\vc o_t,\vc u_t^\star)\) in \(\mathcal B\)
    \FOR{each mini-batch \(\mathcal M\subset\mathcal B\)}
        \STATE Update \(\boldsymbol\pi_\theta\) using \eqref{eq:supervised_loss}
        \STATE Update \(\bar\theta\leftarrow(1-\tau)\bar\theta+\tau\theta\)
    \ENDFOR
    \STATE Evaluate short-horizon hybrid MPC using \(\boldsymbol\pi_{\bar\theta}\)
\ENDFOR
\RETURN trained target policy \(\boldsymbol\pi_{\bar\theta}\)
\end{algorithmic}
\end{algorithm}

\subsection{Distillation Results}

Each controller and planning horizon is evaluated on 160 randomised trials. Success requires spinner-angle
error \(\leq 0.4\,\mathrm{rad}\) for Finger, planar box-to-target
distance \(\leq 0.10\,\mathrm{m}\) for Franka, and base height
\(\geq 0.15\,\mathrm{m}\) together with forward position
\(\geq 1.0\,\mathrm{m}\) for Unitree.

\begin{figure}[ht]
    \centering
    \includegraphics[
        width=\linewidth,
        keepaspectratio=true
    ]{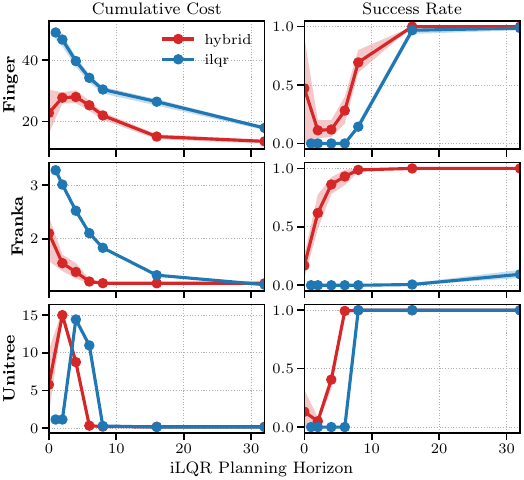}
    \caption{Closed-loop performance versus planning horizon \(H\) at a
\(10\,\mathrm{ms}\) simulation and control timestep over 160 randomised
trials. Left: cumulative episode cost. Right: success rate.}
    \label{fig:learning-horizon}
\end{figure}

\cref{fig:learning-horizon} compares the controllers as the planning horizon varies. For the hybrid controller, \(H=0\) denotes policy-only execution. For \(H>0\), the controller replans at each timestep, applies the first optimised action, and repeats. At \(H=6\), policy guidance improves success by \(28\)--\(98\) percentage points across the three tasks. The policy therefore allows the hybrid controller to succeed with shorter lookahead than standard iLQR.

The largest difference appears on \textbf{Franka}. Standard short-horizon iLQR remains near zero success across the evaluated horizons, whereas the hybrid controller succeeds with substantially less lookahead. The policy supplies the long-horizon pushing strategy, while iLQR corrects the local box contact.

Two task-specific effects further clarify the results. On \textbf{Finger}, the policy-only controller at \(H=0\) can outperform the shortest residual horizons. In this regime, a myopic residual optimiser can perturb a policy that already encodes the required manoeuvre; iLQR becomes beneficial again once its horizon provides sufficient lookahead. On \textbf{Unitree}, very short-horizon control can obtain low cost while remaining near a stationary configuration and avoiding large orientation penalties. Such behaviour does not satisfy the forward-motion criterion, so success is the more informative metric in this regime. The hybrid controller becomes effective once the horizon is long enough to initiate and stabilise forward motion.

\section{Conclusion}
We show that residual-based implicit differentiation makes tolerance-converged MJX contact derivatives practical at accelerator scale. The method matches converged rollout gradients without differentiating through the solver iterations, removing the solver-iteration-dependent memory cost and leaving only the ordinary costs of active contact count and DOF. This enables batched full-horizon iLQR to serve as an online teacher, from which we distil a policy for short-horizon residual MPC. Across Finger, Franka, and Unitree, the resulting hybrid controller achieves substantially higher success than standard iLQR at short planning horizons.

\bibliographystyle{IEEEtran} 
\bibliography{references}

\ifarxiv
\clearpage
\appendices
\section{Training Parameters}
\label{sec::appendix}

We list the full configuration used for each task.

\begin{table}[h]
\centering
\renewcommand{\arraystretch}{1}
\caption{Configuration parameters, Finger task.}
\label{tab:finger_config}
\begin{tabular}{@{}>{\raggedright\arraybackslash}p{0.34\linewidth}
                  >{\raggedright\arraybackslash}p{0.58\linewidth}@{}}
\toprule
\textbf{Parameter} & \textbf{Value / Description} \\
\midrule
iLQR horizon (steps)         & 150 \\
Batch size                   & 128 \\
Epochs                       & 100 \\
Solver iterations            & 10 \\
Control dimension            & 2 (torque control) \\
Observation dimension        & 14 \\
Initial state sampler        & Bimodal on $q_\text{pos}$, uniform on $q_\text{vel}$ \\
Target generator             & Uniform quaternion sampling around Y-axis \\
Mini-batch size              & $(150 \times 128) / 16 = 1200$ \\
Policy optimiser             & AdamW with linear learning rate schedule \\
Policy learning rate         & $1\times10^{-3} \rightarrow 5\times10^{-5}$ over 50 epochs \\
Target increment step        & 10\% of the policy output \\
Network architecture         & $[64, 64]$ \\
Policy norm penalisation     & 0 \\
\midrule
\textbf{Evaluation}          & \textbf{Value / Description} \\
\midrule
Batch size                   & 128 \\
iLQR horizon (steps)         & 6 \\
Solver iterations            & 1 \\
MPC iterations               & 150 \\
\midrule
\textbf{Cost terms}          & \textbf{Description} \\
\midrule
State cost                   & $0.1 (\theta_\text{goal} - \theta_\text{finger})^2 + 5 \times 10^{-4} \|\dot{q}\|^2$ \\
Control cost                 & $0.005 \|u - \pi(x)\|^2$ (running cost) \\
Terminal cost                & Same as state cost \\
Success condition            & $|\theta_\text{goal} - \theta_\text{finger}| \leq 0.4$ \\
\bottomrule
\end{tabular}
\end{table}

\begin{table}
\centering
\renewcommand{\arraystretch}{1}
\caption{Configuration parameters, Unitree A1 task.}
\label{tab:unitree_config}
\begin{tabular}{@{}>{\raggedright\arraybackslash}p{0.34\linewidth}
                  >{\raggedright\arraybackslash}p{0.58\linewidth}@{}}
\toprule
\textbf{Parameter} & \textbf{Value / Description} \\
\midrule
iLQR horizon (steps)         & 200 \\
Batch size                   & 64 \\
Epochs                       & 150 \\
Solver iterations            & 20 \\
Control dimension            & 12 (position control) \\
Observation dimension        & 35 \\
Initial state sampler        & Fixed pose $+$ small velocity noise \\
Target generator             & Fixed unit quaternion (no rotation) \\
Mini-batch size              & $(200 \times 16) / 16 = 200$ \\
Policy optimiser             & AdamW with linear learning rate schedule \\
Policy learning rate         & $1 \times 10^{-3} \rightarrow 1 \times 10^{-5}$ over 50 epochs \\
Target increment step        & 10\% of the policy output \\
Network architecture         & $[32, 32, 32]$ \\
Policy norm penalisation     & 0.15 \\
\midrule
\textbf{Evaluation}          & \textbf{Value / Description} \\
\midrule
Batch size                   & 16 \\
iLQR horizon (steps)         & 6 \\
Solver iterations            & 1 \\
MPC iterations               & 200 \\
\midrule
\textbf{Cost terms}          & \textbf{Description} \\
\midrule
Feet cost                    & $0.005 \sum_i (z_{\text{foot}, i} - 0.07)^2$ \\
Height cost                  & $0.8 (z_{\text{body}} - 0.25)^2$ \\
Linear velocity cost         & $0.005 (v_x - 1)^2$ \\
Rotation cost                & $0.004 \|R(q) - I\|^2$ (deviation from upright) \\
Control cost                 & $0.002 \|u - \pi(x)\|^2$ (running cost) \\
Terminal cost                & Same as height $+$ velocity $+$ rotation terms \\
Termination condition        & $z < 0.15$ or $x < 1.0$ \\
\bottomrule
\end{tabular}
\end{table}

\begin{table}
\centering
\renewcommand{\arraystretch}{1}
\caption{Configuration parameters, Franka arm task.}
\label{tab:franka_config}
\begin{tabular}{@{}>{\raggedright\arraybackslash}p{0.34\linewidth}
                  >{\raggedright\arraybackslash}p{0.58\linewidth}@{}}
\toprule
\textbf{Parameter} & \textbf{Value / Description} \\
\midrule
iLQR horizon (steps)         & 200 \\
Batch size                   & 64 \\
Epochs                       & 100 \\
Solver iterations            & 10 \\
Control dimension            & 7 (torque control) \\
Observation dimension        & 31 \\
Initial state sampler        & Fixed arm qpos; box qpos uniform in $[-0.1, 0.1]\times[-0.2, 0.2]\times\{0\}$ \\
Target generator             & Fixed pose: position $(0.8, 0, 0)$; quaternion from $\pi$ rotation about Z-axis \\
Mini-batch size              & 1200 \\
Policy optimiser             & AdamW with linear learning rate schedule \\
Policy learning rate         & $1\times10^{-3} \rightarrow 5\times10^{-6}$ over 800 steps (50 epochs $\times$ 16 steps) \\
Target increment step        & 10\% of the policy output \\
Network architecture         & $[64, 64]$ \\
Policy norm penalisation     & 0.05 \\
\midrule
\textbf{Evaluation}          & \textbf{Value / Description} \\
\midrule
Batch size                   & 32 \\
iLQR horizon (steps)         & 4 \\
Solver iterations            & 1 \\
MPC iterations               & 200 \\
\midrule
\textbf{Cost terms}          & \textbf{Description} \\
\midrule
State cost, orientation      & $0.001 \|\log(q_{\text{gripper}}^{-1} \otimes q_{\text{target}})\|^2$ \\
State cost, reaching         & $0.05 \|p_{\text{gripper}} - p_{\text{box}}^{\text{offset}}\|^2$ \\
State cost, box to target    & $0.02 \|p_{\text{box}} - p_{\text{target}}\|^2$ \\
State cost, velocity         & $10^{-6} \|\dot{q}\|^2$ \\
Control cost                 & $0.0006 \sum_i R_i (u_i - \pi_i(x))^2$, with $R = [0.4, 1, 0.1, 0.1, 1, 1, 1]$ \\
Terminal cost                & Same as running cost \\
Success condition            & Box-to-target distance $\leq 0.1$ \\
\bottomrule
\end{tabular}
\end{table}

\FloatBarrier

\fi

\end{document}